\documentclass[a4paper]{article}
\pdfoutput=1

\usepackage[ngerman,english]{babel}
\usepackage{xcolor}
\definecolor{fgreen}{rgb}{0.1,0.5,0.2}
\definecolor{grape}{rgb}{0.42,0.2,0.38}
\definecolor{marine}{rgb}{0,0.2,0.6}
\usepackage[hyphens]{url}
\usepackage[bookmarks,bookmarksnumbered,colorlinks=true,linkcolor=grape,citecolor=fgreen,urlcolor=marine]{hyperref}
\definecolor{wmcolor}{rgb}{0.47,0.2,0.3}

\newcommand{\inst}[1]{{$^#1$}}
\newcommand{\email}[1]{\texttt{#1}}
\newcommand{\keywords}[1]{\\[2ex]{\bf Keywords:} #1}
\begin{document}
%
\title{Scope and Sense of Explainability for AI-Systems}
%
%
\author{A.-M. Leventi-Peetz.\inst{1}
  \and T. Östreich\inst{1}
  \and W. Lennartz\inst{2}
  \and K. Weber\inst{2}}
%
%
\date{\small $^1$Federal Office for Information Security, BSI, Bonn, Germany\\
  \email{leventi@bsi.bund.de}
  \\
  $^2$inducto GmbH, Dorfen, Germany\\
  \email{info@inducto.de}}

\maketitle              

\begin{abstract}
Certain aspects of the explainability of AI systems will be critically discussed. This especially with focus on the feasibility of the task of making every AI system explainable. Emphasis will be given to difficulties related to the explainability of highly complex and efficient AI systems which deliver decisions whose explanation defies classical logical schemes of cause and effect. AI systems have provably delivered unintelligible solutions which in retrospect were characterized as ingenious (for example move 37 of the game 2 of AlphaGo).
  It will be elaborated on arguments supporting the notion that if AI-solutions were to be discarded in advance because of their not being thoroughly comprehensible, a great deal of the potentiality of intelligent systems would be wasted.
\keywords{artificial intelligence (AI), machine learning (ML), explainable AI (XAI), chaos, criticality, attractors, echo state networks (ESN), time series, causality}
\end{abstract}

\setcounter{footnote}{0}

\section{Introduction}
The next generation AI-systems are expected to extend into areas that
correspond to human cognition, such as real time contextual events
interpretation and autonomous system adaptation.  AI solutions are
mostly based on \emph{neural networks} (NN) training and inference developed on
deterministic views of events that lack context and commonsense
understanding. Many successful developments have been done in the
direction of explainable AI algorithms while further advancements in
AI will still have to address novel situations and abstraction to
automate ordinary human activities \cite{intelLabsNeuromorphic}.
There exist already various approaches to explain the results of
\emph{machine-learning systems} (ML systems), there are methods and tools
which can interpret and verify for example classification results and
decisions produced on the basis of sophisticated complex ML systems.
The explanations vary with the task and the method
which ML systems employ to reach their results.  The aim of this work
is to give a short but not exhaustive report about known ambiguities,
shortcomings, flaws and even mistakes which ML explainability methods
imply, underlining the association of these problems to the growing
complexity of the systems which have to get interpreted.  Furthermore,
there will be discussed the necessity of taking chaos theoretical
approaches for ML into account, and some implemented
examples which demonstrate the potentiality of this new direction will
be discussed. In conclusion, there will be naturally formulated the
doubt as to whether it is possible, or it makes sense, to follow the
intention of finding ways to make every ML system explainable.  In the
section following this introduction, the importance of making AI
explainable will be emphasized, by reference to some prominent
applications of AI systems, which directly implicate the necessity of
understanding the reasoning behind machine made
decisions. In this context explainability is seen as a requirement of trustworthy
AI. In section~\ref{s:forms}, the difference between the explainability
of rule-based systems of the first generation AI and that of modern ML
systems will be emphasized. Technical aspects of the feature-based
explainability methods for advanced, dynamically adaptive deep learning systems
are discussed in section~\ref{s:dynamical}, with focus on the evaluation of a number of
recent improvements, introduced to increase reliability of explanations.
In section~\ref{s:stability} examples will be given to justify
the comparison of the behavior of ML systems
to the behavior of chaotic systems, whose results are sensitively
dependent on their initial conditions. 
In the following section, advantages of using \emph{echo state networks} and \emph{reservoir computing}
as a computationally efficient and competitive alternative to deep learning methods will
be considered, especially with respect to their ability to simulate both deterministic
but also chaotic systems. In sections~\ref{s:attract} and \ref{s:causal}
the proof of causality in ML results will be presented as an indispensable part of
any sound explanation of ML supported decisions.
At the same time, references will be given to scientific work, which asserts that
the problem of assigning causation in observational data has not yet been solved.
Some AI specialists assign to XAI the property of being \emph{brittle}, easy to fool, unstable or wrong. In the conclusions there will be posed the question if it is absolutely necessary to make all AI systems interpretable in the first place.
At the present state of developments, interpretability does not necessarily contribute to the trustworthiness of AI systems.

\section{Superhuman Abilities of AI}
Of crucial importance is the application of AI in so called safety
and security critical systems, for example in transportation and medicine,
where there is very little or zero tolerance of machine errors.
For instance, the interpretation of ML models employed in \emph{computer-aided diagnosis}
(CADx) to support cancer detection on the basis of digital medical
images is often the recognition of certain patterns which pixels in
the images form \cite{Singh2020ExplainableDeepLearning}.
These patterns are combinations of so called features
(for example gray levels, texture, shapes etc.)  which the algorithm
has extracted from the test image in order to infer a result. The term
inference means ``make a prediction on the basis of experience'', in this
case the experience which the model has gained during its training
phase, exploiting information stored in large labeled datasets. This
would be the case of supervised learning which tought the model to
discern between pathological and normal forms.
The increasing accuracy of imaging methods calls for an increase of accuracy 
and reliability of the algorithmic predictive mechanisms.
Imaging examination has no longer only qualitative and pure diagnostic
character, it now also provides quantitative information on disease
severity, as well as identification of so called biomarkers of
prognosis and treatment response. ML systems are committed with the
objective of complementing diagnostic imaging and helping the
therapeutic decision-making process. There is a move toward the rapid
expansion of the use of ML tools and leading radiology in daily life
of physicians, making each patient unique, in the concept of
multidisciplinary approach and precision medicine.
The move from the well established predictive analysis to the so
called prescriptive one, one that should expect systems to be even
more efficient and in a way smarter gets stronger. The quality of
these systems concerns not only the health sector but also industry
and economy, regarding for example the emergence of smart factories
and the approaching realization of the fourth industrial revolution
(\emph{industry 4.0}) with the planning of self-organizing intelligent
systems, that is systems which can anticipate and find solutions for
suddenly arising problems, and most probably also unforeseen problems
by themselves. This new generation of system automations will
probably have an enormous social and economic impact world
wide. People, societies, will have to rely on the decisions and the
advices of machines to organize life. But can advices and decisions of
machine systems get completely trusted eventually even without the
final approval of some reviewing human experts committee? Could they
be accounted as reliable and secure? Could people perhaps trust these
systems if their behavior becomes somehow explainable? In this case
could the development of adequate norms and criteria as to how machine
explanations should look like be enough in order to inspire trust?
And who should be able to comprehend these explanations?
These questions have received a great deal of attention in
the last years and will stay in focus of research for a long time to
go.  
Explainability has received special attention ever since AI algorithms
managed to reach what is being called \emph{superhuman} abilities.
People have realized that they can develop systems that are not only
faster in solving problems but can also do better, because they can
find solutions which no expert has ever been able to find so far.  One
has to recall the famous \emph{creative} and \emph{unique} stone move
37 of the game 2 of AlphaGo which was evaluated by AlphaGo as having a
probability of being played by a human close to one in ten million \cite{canaan2019playingfield}.
Experts have been asked about the implications of this kind of
creativity. Some of the experts attributed the move to clever
programming, and not creativity of the software. In other articles the
advancements from AlphaGo to AlphaGo Zero ( a program that can win a
play without any use of information based on human experience) has been
seen as an
example of the AI becoming self-aware and creating its own AI which is
as smart as itself if not smarter.  Experience shows that experts in
general cannot always make explanations of their decisions
understandable not even for fellow experts! However it is expected
that the self-awareness of AI systems should enable them to explain
their decisions to humans.  In fact on AI systems there are made much
higher demands than on humans when they have to make decisions. In
autonomous driving for example it is expected that the technology must be
at least 100 times better than humans, according to Prof. Trapp of
Fraunhofer ESK \cite{Strehlitz2019InterviewTrapp}.

\section{Forms of Explainability}
\label{s:forms}
The rule-based systems, or expert systems of the first generation of AI,
were deterministic. Their intelligence was fixed, following a
definite series of rules and instructions, their inference was made
based on boolean or classical logic. The explanation of the decisions of
those systems was the demonstration of the inference rules that led to
a decision. But these systems followed rules which would be determined by
humans. They were as causal, fair, robust, trustworthy and usable as
their developers had made them to be.  These systems wouldn't change or update on
their own, they would not learn from mistakes. They simulate AI but
for many experts they were not true AI systems.

The first so called \emph{reason tracing explanations} were saying
nothing about the system’s general goals or resolution strategy
\cite{david1993ExpertSystems}.  The utilization of the fact that
knowledge of the problem to be solved, if expressed in a form that
computers can handle, offered advantages, motivated domain experts, so
named knowledge engineers, to encode experts' advice in the form of
associational (also referred to as heuristic or empirical) rules that
mapped observable features (evidence) to conclusions. For a large
portion of real-world problems it is significantly easier to collect
data and identify a desirable behavior, than to explicitly write a
program, as Karpathy aptly stated (2017) \cite{Karpathy2017SW}. ML systems, nowadays powered by NN and deep
learning shifted the paradigm from one in which the programmer must
provide rules and inputs in order to obtain results, to one where
specialists and no specialists can provide inputs and results to
derive rules. The promise of this approach is that learned rules can
be applied to many new inputs, without requiring that the user has the
expertise needed to derive results. This is sometimes also observed as
democratization of AI.

The motivation in this respect is that representing knowledge in datasets is much
easier than having to provide methods of encoding and manipulating
symbolic knowledge. Because in this case updating and improving
learning systems can be done more smoothly as the datasets grow and
evolve over time. Furthermore, rule-based systems are not of help for
solving problems in complex domains and there are many cases (e.g.,
cancer detection in medical images), where no
explicitly defined rules in a programmatic or declarative way are possible.

The hope of AI research is to implement general AI by
creating autonomously learning systems.
These systems should become
finally unlimited in their ability to simulate intelligence, they should
be able to demonstrate all signs of an adaptively growing intelligence:
Previous knowledge should be modified, eliminated if not needed any more,
while new knowledge should be continuously
gained. Hence, these systems should be able to build and update their rules actively
on the fly. This is the difference
between ML systems and rule-based ones. 
Neural networks represent instances of learning systems.
A learning system implements a utility function representing the
difference between the system's prediction and reality and this
difference will be minimized for example with the help of optimization
techniques, which will change the system's parameters.
These optimization techniques (e.g., gradient descent, stochastic
gradient descent) are in fact rule-based techniques because 
they just compute gradients needed to adjust the
weights and biases to optimize its utility
function. The approach of the calculation varies considerably (e.\,g.,
between supervised and unsupervised learning).
The learning process is deterministic (including the statistical
and probabilistic part of the method), however it is practically impossible
to describe the
learning system with a model because this would involve millions
of dynamic parameters (e.\,g., weights, biases) which make the description
of internal system processes untraceable.
Their enormous complexity makes learned systems very hard to
explain, so that they can hardly get understood by humans \cite{TricentisAIApproaches}.

It can't become entirely clear for trained systems how they make their
decisions. That’s the dark secret at the heart of learning systems
according to Will Knight, Senior Editor of MIT Technology Review.\footnote{\url{https://www.technologyreview.com/author/will-knight/}}
According to Tommi Jaakkola (MIT, Computer Science)\footnote{\url{https://people.csail.mit.edu/tommi/}} this is already a major problem for many
applications; whether it’s an investment decision, a medical decision,
or a military decision, one doesn't want to just rely on a black box.

The European Union issued the so named EU General Data Protection
Regulation \cite{europeGDPR}
which is practically a right to explanation. Citizens are entitled
to ask for an explanation about algorithmic decisions made
about them. There arises the question if GDPR will become a game-changer for
AI technologies. The consequences of this regulation are not yet
really clear. It remains to be seen whether such a law is legally
enforceable. It’s not clear if that law is more a right to inform
rather than a right to explanation. Therefore, the impact of GDPR on
AI is still under dispute.
For the explainability of NN models, a
large body of work focuses on post-training feature visualization to
qualitatively understand the dynamics of the NNs.
The following properties are important for explanations:
\begin{itemize}
\item Causality
\item Fairness
\item Robustness and Reliability
\item Usability
\item Trust
\end{itemize}
There have been long discussions about biased decisions, the famous
husky which has been misidentified as wolf, because of the snow in the
picture of his environment, is known to almost everybody. The bias in
the data is a serious issue especially because as experts point out,
algorithms tend to amplify existing biases, they actually learn from
differences and any difference can under circumstances become a bias
in the process of learning. However one cannot discard the possibility
that even if all training datasets were balanced, so that no biases
were possible, there could always still exist some kind of biases in
the opinion of users who are meant to understand the algorithm's
interpretation and judge about the algorithmic fairness.
There are many subtleties involved in interpretation which should be
of concern in parallel to the technical refinement of algorithms and 
software.

\section{Complex Dynamical Systems}
\label{s:dynamical}
Learning setups can not always be static.
The necessity of learning in continuous time, by using continuous data
streams to which also online learning belongs, has established
incremental learning strategies to account for situations
that training data become available in a sequential order.
 The best predictor for future data gets updated at each step, as opposed
to batch learning techniques which generate the best predictor by
learning on the entire training data set at once.
Online learning algorithms are also known to be prone to the so named
catastrophic interference, which is the tendency of an artificial
NN to completely and abruptly forget previously learned
information upon learning new. This is the well-known
stability-plasticity dilemma~\cite{catastrophicInfe}.

An algorithm has to dynamically adapt to new patterns in the data,
when the data itself is generated as a function of time, e.g., stock
price prediction. In time series forecasting a model is employed to
predict future values based on a previously performed time series
analysis and the thereof values observed. That is historical
data is used to forecast the future. Such predictions are delivered
together with \emph{confidence intervals} (CI) that reflect the confidence
level for the prediction. The size of the sample and its variability
belong to the factors which affect the width of the confidence
interval, as well as the confidence level, usually set at 95\,\% \cite{Brownlee2019ConfInt}.
A larger sample will tend to produce a better estimate of the population
parameter, when all other factors stay unchanged.  However, NNs belonging to specific settings do not provide a unique
solution, because their performance is determined by several factors,
such as the initial values, usually chosen randomly from a distribution, the
order of input data during the training cycle and the number of
training cycles \cite{GrossiParametersTimeSeries,RichterLSTMblog,Cui2016ContOnlineSeqL}.
Other variables belonging to the
mathematical attributes of a specific NN, like learning rate,
momentum, affect also the
final state of a trained NN which makes a high number of
different possible combinations possible. Evolutionary algorithms have
been proposed to find the most suitable design of NNs, in
order to allow a better prediction, given the high number of possible
combinations of parameters. Also many different NNs can be
trained independently with the same set of data, so that an ensemble of
artificial NNs that have a similar average performance but
a different predisposition to make mistakes on their individual level
of prediction will be created \cite{CerlianiMarcoNetworksEnsembles}.
If one needs to estimate a new patient's individual risk, for example in
cardiovascular disease prediction, or the riskiness of a single stock,
or one must classify the danger of some unknown
data traffic pattern that might hide a cyber attack,
a set of independent NN models acting simultaneously on the same
problem should be of advantage. An ensemble of models performs
better than any individual model, because the various errors of the
models \emph{average out} therefore it has dominated recent ML competitions \cite{CharuMakhijaniEnsembleLearning,GarbinDropout2020}.
Using model ensembles also requires a much larger training time as compared
to training only one model. Each model is trained either from scratch
or derived from a base model in significant ways. In all kinds of
ensemble methods, concatenated, averaged, weighted etc., one has
certain advantages and disadvantages and a reported accuracy of up to
89\,\% on test data. Explainability refers to the ability of a model or
an ensemble of models to explain its decisions in terms of human
observable decision boundaries or features.
Should the user get a proof that a different choice of ensemble
weights would not have resulted to a different classification in his
case? How do the decision boundaries look like that resulted to the decision
concerning him? One can also develop ensembles during fine-tuning
operations dividing the procedure in subtasks.

Incremental and active learning remain a field of
research aiming at developing recognition and decision systems that are able to
deal with new data from known or even completely new classes by
performing learning in a continuous fashion. Active learning and
active knowledge discovery are approaches, which require continuously
changing models. How should continuous learning with a series of
update steps get performed robustly and efficiently is a question
that still remains open. And how explainable are these models for the user?

If it is allowed to assume that the parameters of the NN vary
smoothly with the time-varying training dataset, one can apply warm-start
optimization for each time step, using the parameters of the previous
step as initialization for the current parameters. In this case a
network fine-tuning is performed under the assumption that the
introduction of new categories is not necessary for the classification
of the new data. If however the new datasets have little or nothing in
common with the datasets of the previous step, new classes (known or
unknown) have to be added with additional nodes at the output layer of
the network, together with some new parameters and a new normalization
for this network. Questions of convergence under time limitation or
perhaps data sparsity are in general open. How many layers must be
adapted so that a robust solution can be found for real-world and
real-time applications. For example how many SGD (\emph{stochastic gradient
decent}) iterations would be necessary
for each update in order to achieve calculation accuracy without the need of
overwhelming computational effort. There exist empirical studies
which have investigated various factors among others the fraction of
older to new data to be considered during the SGD iterations as to avoid overfitting. 
The dropout technique randomly changes the network architecture to minimize the risks that learned parameters do not generalize well.
This method in essence simulates ensembles of models without creating multiple networks.
The dropout technique requires tuning of hyperparameters to work well, like change
the learning rate, weight decay, momentum, max-norm, number of units
in a layer, and for a given network architecture and input data
requires experimentation with the hyperparameters. Dropout increases
convergence time as one needs to train models with different
combinations of hyperparameters that affect model behavior, further
increasing training time \cite{GarbinDropout2020}. However dropout acts detrimental to accuracy
if used without normalization therefore normalization techniques have
been developed, some also going beyond the batch normalization to
account for active learning.
On top of this, wrong object labels (label noise) are not completely
avoidable in real-world applications which considerably degrades the
accuracy of the results. Researchers have managed to spot changes of
a continuously learning deep CNN (\emph{convolutional neural network}) by visualizing the shifting of the
mainly attended image regions, for example when a new class is
introduced, by observing the strongest network-filter changes during a single
learning step \cite{Bollt2018CausationInferenceDynamicalSystems}.

Visual explanations for DNN---for example CAM (\emph{class activation mapping}) or Grad-CAM~\cite{BuhrmesterGradCAM2019,Montavon2019LRP}---are
posthoc, they work on a NN after training is complete and
the parameters are fixed, when also for only a short time. The network
produces a feature map at its last convolution layer, and weights of
features or gradients with respect to feature map activations are
posthoc calculated and plotted. The result is a class-discriminative
localization map which determines the position of particular class
objects. However explainability is not interpretability and therefore
posthoc attention mechanisms, although perhaps helpful for following reactions
of agents in video games, may not be optimal for real-world decisions
connected with high risk.
Explaining how a model made its decision delivers a chain of
results, after a sequence of mathematical operations have been applied
to the model and can perhaps help to better understand the functionality of the
model but it does not also provide any known rules of the
natural world which would make sense to humans.
Moreover, model rules do not always translate to unique or
comparable decisions, so that to find a way to
translate model rules (explainability) to natural world rules
(interpretability for humans) would not be the only problem that 
has to be solved.
For instance studies have demonstrated that the overlap of features,
which filters extract in high convolutional layers, leads to poor model
expressiveness in CNNs. Methods have been developed to remove
redundancies and feature ambiguity by inducing bias in the training
process and confine each filter's attention to just one or a few
classes.
Also methods to disentangle middle-layer representations of CNNs to
correspond to objects and to object parts features have been developed,
in order to assign semantic meanings to filters \cite{zhangInterpretableCNN}. Because there is a
trade-off between explainability and performance, in real-world
applications additional networks, so called explainability networks,
have been implemented and trained in parallel to the original
performing networks with the task to make the former explainable. For
the training and testing of explainable filters, benchmark datasets
with \emph{ground truth} annotations have been employed. In a number of cases
the majority of classifications could be attributed to these new
filters, but there have been also cases where the performing network
achieved better classifications than its corresponding explaining
network.
The additional computational effort and time associated with the
process of features disentanglement makes the concept not
applicable for dense networks or when a great
number of features have to be recognized \cite{Raffin2019DecouplingFeatExtr}.
CNNs use pooling which is the application of down sampling of the
feature map to ensure that the CNN recognizes the same object in
images of different forms and also to reduce the memory requirements
of the model. The pooling operation introduces spatial invariance in
CNNs which is also one of the major weaknesses of CNNs.  Max pooling
for instance preserves the best features and the feature map gets
flattened into a column matrix to be processed in the NN for further
computations. As a result of pooling, CNNs can lose features in images
and there would be needed a very big amount of training data for 
this weakness to get
compensated. CNNs are also unable to recognize pose, texture and
deformations in images or parts of images. CNNs lack equivalence
because they don't implement equivariance, however they use
translational invariance therefore they can for example detect a face
in a picture, if they have detected an eye, independent of the spatial
location of the eye in respect to the rest of features which usually
belong to a face.

Alone on the basis of features the results of a CNN cannot generally
get interpreted as it seems. Capsule networks or CapsNets have been
proposed as an alternative to CNNs \cite{MensahCapsuleNetworks2019}.
Their neurons accept and output
vectors as opposed to CNNs’ scalar values.  Features can be learned
together with their deformations and viewing conditions.  In capsule
networks, each capsule is made up of a group of neurons with each
neuron’s output representing a different property of the same feature.
The output of a capsule is the probability that a feature is present
and is delivered together with the so named instantiation parameters,
expressing the equivariance of the network, or its ability to keep its
decision unchanged regardless of input transformations.

The introduction of CapsNets is considered to be promising for solving
real life problems like machine translation, intent detection, mood
and emotion detection, traffic prediction on the basis of
spatio-temporal traffic data expressed in images etc.  Even though the
training time for CapsNets is better than CNNs, it is still not
acceptable for time critical operations and highly unsuitable for
online training. Research is currently ongoing in this area.
CapsNets are considered to be explainable by design, because during
learning they construct relevance paths that reduce unrelated capsules
without the necessity of a backward process for explanation.

\section{Stability and Chaos}
\label{s:stability}
An important issue concerning the trustworthiness of DNNs is their
liability to mistakes when adversarial examples are introduced as
inputs to them causing them wrong decisions. Intentionally designed
examples to fool a model, are the adversarial attacks, which some call
optical illusions for machines, as they mostly concern widely
discussed examples of striking miscategorization of pictures. Quite
famous is the case of the classification network which had been
trained to distinguish between a number of image categories with panda
and gibbon being two of them. The classifier determines with 57.7\,\% of
accuracy the image of a panda. If a small perturbation is added to the
picture, the classifier classifies the image as gibbon with 99\,\%
accuracy \cite{Goodfellow2015ExplainHarnessAdversarial}. Research has showed that the output of \emph{deep neural networks} (DNN) can be easily changed by adding relatively small
perturbations to the input vector. There exist also designed and
successfully applied attacks with an one-pixel image perturbation, for example
based on what is called \emph{differential evolution} (DE) which can fool more
types of networks \cite{SuOnePixelAttack2019}.
\emph{Reinforcement learning} (RL) is the autonomous learning of agents who
learn out of experience how to carry out a designated task, and
discover the best policy of behavior, or the best actions to undertake
through interaction with their environment and evaluation of the
according collection of rewards and punishments. RL systems have been proved
to be also liable to mistakes due to adversarial attacks.
It has been demonstrated that learning agents can also be manipulated
by adversarial examples. Research shows that widely-used RL
algorithms, such as DQN (\emph{deep Q-learning}), TRPO (\emph{trust region policy optimization}), and A3C (\emph{asynchronous advantage actor critic}), are vulnerable to adversarial
inputs. Degraded performance even in the presence of perturbations which are too
subtle to be perceived by a human, can cause an agent to make wrong
decisions \cite{ChenXiangAdversarialRL2019,Ilahi2020AdversarialAtt}.
ML systems are highly complex and complexity makes a
system itself highly dependent on initial conditions. The here
mentioned examples, where a small perturbation causes the system to
make a jump in category space, present an analogy to the behavior known of
chaotic systems, small changes in the starting state can generate a
big difference in the dynamics of the system later on.
The noise needed to add to the panda picture in order 
to get the false classification was a so named custom made perturbation,
especially generated by a GAN, a \emph{generative adversarial network}, trained to 
fool models by exploiting chaos.
Perturbations can be meticulously designed to serve certain purposes,
and make a DNN take a wrong decision, however also completely random perturbations
which can arise accidentally in very complex environments where ML is
already applied or is planned to be applied in the near future,
especially implicating systems with real time requirements, can cause
serious mistakes with possibly catastrophic consequences. In certain
cases it can be difficult to discern between input signal and perturbation.
There is a close relationship between complex systems research and
ML with a wide range of cross-disciplinary interactions.
Exploring how ML works in the aspect of
involving complexity is a subject of significant research which has to
be considered also in the context of interpretation \cite{Tang2020NetworksChaos}.
For \emph{time series classification problems} (TSCP) features have to
get ordered by time, unlike the traditional classification
problems. CNNs have been applied on time series automatically,
tailoring filters that represented repeated patterns, learned and
extracted features from the raw data.
\emph{Recurrent neural networks} (RNNs) are a family of NN used
especially to address tasks which involve time series as input, and are
therefore deployed in sequential data processing and continuous-time
environments. They are capable of memorizing historic inputs, they
posses dynamic memory, as they preserve in their internal state a
nonlinear transformation of the input history. They are characterized
by the presence of feedback connections in their hidden layer which
allows them short-term memory capability. However their learning of
short and long-time dependencies is problematic when implemented by
means of gradient descent (vanishing/exploding gradients) whereby
their training with backpropagation through time is computationally
intensive and often inefficient.
The interpretability of the internal dynamics of RNNs is input
dependent and almost infeasible given the complexity of
the time and space dependent activity of their neurons.

\section{Nonclassical Approaches, Training of Attractors}
\label{s:attract}
Nonclassical approaches like for example some based on heteroclinic
networks with multiple saddle fixed points as nodes, connected by
heteroclinic orbits as edges in the phase space of the learning system
have been elaborated to generate reproducible sequential series of
metastable states and attractors to explain RNN behaviors. To this
task, known engineering methods have been extended to enable data
based inference of heteroclinic dynamics \cite{VoigtInferenceEteroclinicNet}

These approaches use \emph{reservoir computing} (RC)
and reservoirs, that can be employed instead of temporal kernel functions,
to avoid training-related challenges associated with RNN (slow convergence
and instabilities etc.).
\emph{Echo state networks} (ESNs) and \emph{liquid state machines} (LSM)  have been
proposed as possible RNN alternatives, under the name of 
RC. Reservoirs, seen as generalizations of
RNN-architectures and ESNs, are far easier to train 
and have been mainly associated with supervised learning 
underlying RNNs. 
They map input signals into higher dimensional computational spaces
through the dynamics of fixed, non-linear systems, the
reservoirs. ESNs are considered appropriate to be used as universal approximators
of arbitrary dynamic systems. Furthermore, the NN of the reservoir
is randomly generated and only the readout has to be trained. The
trained output layer delivers linear combinations of the internal
states, interpreting the dynamics of the reservoir and its
perturbations by external inputs. Reservoir computing can be applied for
model-free and data based predictions of nonlinear dynamic systems.
Reservoirs can be also applied for continuous physical
systems in space and/or time, allowing computations in situations
where partially or completely unknown interactions or extreme
variations of the input signal take place, allowing for very limited
functional control and almost no predictability.

Andrea Ceni, Peter Ashwin, and Lorenzo Livi have investigated the
possibility to exploit transient dynamical regimes and what they
define as excitable network connections to switch between different
stable attractors of the model for classification purposes~\cite{ceniExcitableAttractors2020}.
They demonstrated how to extract such \emph{excitable network attractors}
(ENAs) from ESNs, whereby the previous training induced bifurcations
that generated fixed points in phase space so that the trained system
under small perturbations as input could move from one stable
attractor to another. The hope is, that this can get exploited for classification
problems that involve switching between a finite set of classes
(attractors) and could be used instead of RNNs. Input dependent
excitability thresholds of excitable connections have been also
defined to measure the minimum distance in phase space, which would be necessary in
order for a solution to escape from a stable point and converge to
another.
The authors found out that there exist \emph{local switching subspaces} (LSS)
in the vicinity of attractors, the dimensions of which directly relate
with the activity of connections in the network, when the ESN solves a
task, in dependence of the complexity of the input and its impact on
the dynamics of the reservoir. And this has to be assessed on a
case-by-case basis. Finding fixed points for the dynamics of the
system depends on the convergence of the optimization algorithm and
one can have similar solutions, which in dependence of the chosen
tolerance can be numerically different. Excitability thresholds should
be important for the robustness of the solutions. ESNs which yield network attractors
with low excitability thresholds were found to be less robust to noise
perturbations. But sensitivity and accuracy of the network do suffer
under low excitability.
Training of the reservoir is simply tuning the readout parameters using comparison
between input and output data, and an autoregressive process to minimize the
difference.
The result of the training could be for example a
classification system which, when a sequence of patterns is given, can
recognize each pattern by itself. A trained reservoir should act as an
autonomous dynamical system whose state evolution, given the initial
conditions, represents the state evolution of the nonlinear dynamical
system that has to be predicted (task system). The forecast horizon
is used to estimate the quality of short-time predictions of such a
trained system. It is defined as the time between the start of a
prediction and the point where it deviates from the test data more
than a fixed threshold.

There have been investigations, as to how to choose training
hyperparameters like reservoir size, spectral radius, network
connectivity, training sample size, training window and so on, in
order to get reliable predictions. The latter must compare to the
typical time scales of the motion of the system, determined by the
maximum Lyapunov exponent. However the calculation of the Lyapunov
exponent is complex and numerically unstable and one needs to have a
knowledge of the mathematical model of the system to calculate it.
This is not the case if one has only the time series data.
The dynamics of a system can also be multiscale, noisy which might
sometimes lead to rare transition events. Some systems can also spend
very long periods of time in various metastable states and rarely, and
at apparently random times, due to some influencing signal, suddenly
transform into a new, quantitatively different state. Such changes in
the dynamical behavior of complex systems are also known as critical
transitions and occur at so-called tipping points. Theories explain
this behavior as due to a large separation of time scales between the
system state and signal evolve. Also complex and multiscale data have
to be analyzed for system behavior predictions.

It is an open question, how good can events and also rare events get
predicted in multiscale nonlinear dynamic systems, making use of only
the slow system state data for the training and having perhaps only a partial knowledge of
the physics of the data generating system.
In this context there exist developments in the direction of what
is called physics-informed ESNs, which are ESNs extended to represent
solutions of ODE (\emph{ordinary differential equation system}), aiming at introducing causality in ML.
Physical information gets imposed in the reservoir by means of special
constraints of invariant principles. The ESN-architecture should be
represented by an ODE approximator, which implements a physics-informed
training scheme for the reservoir computing model \cite{doanPhysicsESN}.
Jiang et al. (2019)~\cite{jiangESNspectralRad}
have demonstrated for reservoir computing systems
which were employed for model-free prediction of nonlinear dynamical
systems, that there exists an interval for their spectral radius within
which the prediction error is minimized. The authors have performed
many experiments keeping the many hyperparameters of the reservoir fixed
and leaving only the edge weights free. Characteristic for a reservoir consisting of a
complex network of $N$ interconnected neurons, is its
adjacency matrix, an $N\times N$ weighted matrix, whose largest absolute
eigenvalue is the network's spectral radius.
The authors have used ensemble-averaged predictions to show that the
spectral radius of the reservoir plays a fundamental
role in achieving correct predictions.
They substantiated this finding
by experimenting with a number of spatiotemporal dynamical systems
known from physics: the \emph{nonlinear Schrödinger equation} (NLSE), the \emph{Kuramoto Sivashinsky equation} (KSE), and the \emph{Ginzburg-Landau equation} (GLE), where they could
compare between the evolution of the true solution with the according
results delivered by trained ESNs. For all the examined systems there
could be found optimal intervals for the values of the spectral radius,
and it could be determined that, when the radius lies outside this interval the prediction error raises immensely.
This result remained valid, independent
of the rest of the network parameters.
Also in a case where performed calculations showed that only about 50
out of 100 ensemble realizations resulted to acceptable predictions,
the spectral radius still had to be taken out of the optimal interval in order
to get reasonable results in terms of accuracy and time.
Remarkable is that also in the case of
a chaotic nature of the solution, the necessity of
choosing the spectral radius out of the optimal interval in order to get
a meaningful predictions remains valid. Furthermore, it could be
demonstrated that using directed or undirected network topology
strongly influences the magnitude of the spectral radius interval, the
directed case leading to different spectral radius values and also to an absolute
minimum of achievable prediction error~\cite{jiangESNspectralRad}. 
While traditional methods for chaotic dynamical systems manage to
make short-term predictions for about one Lyapunov time,
model-free reservoir-computing predictions based only on data demonstrate 
a prediction horizon up to about half a dozen Lyapunov times~\cite{jiangESNspectralRad}.
It has also been discovered that the computational efficiency of ESNs gets
maximized when the network is at the border between a stable and an
unstable dynamical regime, at the so called edge of criticality or the region at the edge of chaos.
That makes especially interesting 
the state between ordered dynamics (where disturbances die
out fast) and chaotic dynamics (where disturbances get amplified). The
average sensitivity to perturbations of its initial conditions allows
to decide if a dynamical system has ordered or chaotic dynamics.
There seems to exist no standard recipe of how to design an RNN or an ESN so that
it operates steadily at its critical regime independent of task
properties. Researchers suggest the development of mechanisms for
self-organized criticality in ESNs \cite{verzelliCriticality2019,pathakAttractors2017}.
Could a guarantee for a very low error in results, finally substitute the demand
for explanations of ML systems predictions, so as to categorize them as trustworthy, without case-dependent technicalities, like counterfactual explanations, feature-based explanations, adversarial perturbation-based explanations etc.
It is quite obvious that using established XAI methods, the creation of explanations would find it difficult to keep pace with the rate of production of results that need to be explained (dynamical systems, online learning, IIoT etc.).
\section{Causality of results?}
\label{s:causal}
It is plausible to consider that it is difficult to have trust or a
comprehensible interpretation of the results of ML
and deep learning, unless
causality regarding the production of these results can be established
as a basis for the interpretation. Causality implicates temporal
notion in the sense that there is a direction in time which
dictates how a past causal event in a variable produces a future event
in some other variable, which leads to a natural spatiotemporal
definition of causal effects, that can be used to detect arrows of
influence in real-world systems \cite{BiancoMartinez2018SpaceTimeCausa}.  
Mechanistic models which get fitted to predict results in
complicated dynamical systems, represent simplified versatile
descriptions of scientific hypotheses, and they implement
parameters which are interpretable as they
have a correspondence in the physical world. It is different with
causation inference from data, the so named observational inference,
the causality of which constitutes a challenging problem for complex
dynamical systems, from theoretical foundations to practical
computational issues \cite{Bollt2018CausationInferenceDynamicalSystems}.
Granger's causality formulation
describes a form of influence on predictability (or the lack of
predictability), in the sense that from time dependent observations of a
free complex system, without any probing activities exercised on it, it examines if
the knowledge of one time series is useful in forecasting another time
series, in which case the former can be seen and interpreted as
potentially \emph{causal} for the latter.

The question of causation is fundamental for problems of control,
policy decisions and forecasts and there can be probably no decision
explanation without revealing the causation inference of the decision
supporting system.
Measures based on the Shannon entropy informational-theoretic approach,
allow for a very general characterization of
dependencies in complex and dynamical systems from symbolic to
continuous descriptions.
In analogy to Wiener-Granger causality for
linear systems,
transfer entropy is a way to
consider questions of pairwise information transfer between nonlinear
dynamical systems. 
However several works have shown limitations in measuring dependence and causation.
Some researchers examine the causation problem with respect to dynamical attractors and the concept of generalized synchronization.
Convergent cross-mapping tests implement the examination of the 
so named closeness principle.
Within the framework of \emph{structural causal models} (SCMs) there have been
examined conditions under which nonlinear models can be identified
from observational data. This method does not always deliver unique
solutions however.

\section{Conclusions}
ML algorithms and their implementations are inherently
highly complex systems and the quality of their predictions under
real-world operation conditions cannot be safely quantified. To
explain the functionality of a deep-learning system under the
influence of an arbitrary input of the domain for which the system
has been designed and trained for, is considered to be generally impossible.
NN based ML systems will be explained mostly through observations of the
magnitude of network activations along paths connecting their neurons, followed back to the network input. Especially popular are XAI visualizations for interpretability,
which highlight those parts of an image which are mostly correlated to the classification
result (attention-based explanations).
Such explanations are not always unambiguous,
they are not intuitive, repeatable or unique. Arun Das et al. (2020)~\cite{das2020XAISurvey} write about the ``inability
of human-attention to deduce XAI explanation maps for decision-making and the unavailability of a quantitative measure of completeness and correctness of explanation maps''. 
The authors recommend further developments, if visualization techniques should be used
for mission-critical AI applications. Returning to causality,
it has to be emphasized, that causal inference from observational data
is an open issue and still a subject of research. Attempts to create explainable
surrogate models, for example using ODE systems (for instance neural
ODE architectures for sequential data processing) adapting the
equations parameters with the help of ML, underlie
uncertainty and errors.  Could dynamical systems get endowed with some
kind of \emph{self-awareness}, that is could they manage to
maintain an inbuilt mechanism of internal active control, able to
instantaneously evaluate the system's state, if it is ordered,
critical or chaotic, this would empower them to even ask for human intervention. However, the time scale on which
systems undergo phase transformations and the duration of their stay
in new states are beyond control, so that a request might have lost
actuality, before a human specialist can react, let alone the
possibility to prevent undesired system decisions, by forcing some
alternative decision or even stopping the system.  Such an option
would be a contradiction in itself because AI
systems are developed and employed to produce decisions correctly and
fast based on data alone, as they are intended for tasks which
no human experts can efficiently perform. This accounts of course for
the cases when the AI systems operate as desired by their developers.

Another matter is the significance and the priority of explanations,
for example when a new, unforeseen and therefore not assessable
decision has been delivered.  Getting back to the \emph{creative} and
\emph{unique} move 37 of the game 2 of AlphaGo, which would have been
chosen with probability close to one in ten million, how could it have
ever been possible to explain this move to someone and convince him in
advance that this is indeed the right move to make in order to win the
game?  The tendency goes to a growing need for creative and unique
decisions generated by AI systems for a world of increasing
complexity, to open the way to new perceptions and novel concepts.
For example, could AI prevent a disaster by timely predicting
unforeseen threats?  In this sense many AI systems may have to stay
unpredictable to deal with unpredictable and even chaotic
circumstances, which call for unexpected solutions inherently lacking explanations,
that build upon previous experience and already discovered knowledge.

%


\begin{thebibliography}{99}


\bibitem{BiancoMartinez2018SpaceTimeCausa}
  Ezequiel Bianco-Martinez, Murilo S. Baptista: Space-time nature of causality. Chaos 28, 075509 (2018). \url{https://doi.org/10.1063/1.5019917}
  
\bibitem{Bollt2018CausationInferenceDynamicalSystems}
  Erik M. Bollt, Jie Sun, Jakob Runge: Introduction to Focus Issue: Causation inference and information flow in dynamical systems: Theory and applications. Chaos 28, 075201 (2018). \url{https://doi.org/10.1063/1.5046848}

\bibitem{BuhrmesterGradCAM2019}
  Vanessa Buhrmester, David Münch, and Michael Arens: Analysis of Explainers of Black Box Deep Neural Networks for Computer Vision: A Survey. arXiv e-print (2019). \url{https://arxiv.org/abs/1911.12116}
  
\bibitem{Brownlee2019ConfInt}
  Jason Brownlee: Confidence Intervals for Machine Learning. Tutorial at Machine Learning Mastery (2019). \url{https://machinelearningmastery.com/confidence-intervals-for-machine-learning/}

\bibitem{canaan2019playingfield}
  Rodrigo Canaan, Christoph Salge, Julian Togelius, Andy Nealen: Leveling the Playing Field -- Fairness in AI Versus Human Game Benchmarks. arXiv e-print (2019). \url{https://arxiv.org/abs/1903.07008}

\bibitem{ceniExcitableAttractors2020}
  Ceni A., Ashwin P., Livi L.: Interpreting Recurrent Neural Networks Behaviour via Excitable Network Attractors. Cogn Comput 12, 330–-356 (2020). \url{https://doi.org/10.1007/s12559-019-09634-2}

\bibitem{CerlianiMarcoNetworksEnsembles}
  Marco Cerliani: Neural Networks Ensemble. Posted on towards data science (2020). \url{https://towardsdatascience.com/neural-networks-ensemble-33f33bea7df3}

\bibitem{CharuMakhijaniEnsembleLearning}
  Charu Makhijani: Advanced Ensemble Learning Techniques.  Posted on towards data science (2020). \url{https://towardsdatascience.com/advanced-ensemble-learning-techniques-bf755e38cbfb}

\bibitem{ChenXiangAdversarialRL2019}
  Chen, T. et al.: Adversarial attack and defense in reinforcement learning-from AI security view. Cybersecur 2, 11 (2019). \url{https://doi.org/10.1186/s42400-019-0027-x}

\bibitem{Cui2016ContOnlineSeqL}
  Yuwei Cui, Subutai Ahmad, Jeff Hawkins: Continuous Online Sequence Learning with an Unsupervised Neural Network Model. Neural Computation 28, 2474-–2504 (2016). \url{https://numenta.com/neuroscience-research/research-publications/papers/continuous-online-sequence-learning-with-an-unsupervised-neural-network-model/}

\bibitem{das2020XAISurvey}
  Das A., Rad P.: Opportunities and Challenges in Explainable Artificial Intelligence (XAI): A Survey. arXiv e-print (2020). \url{https://arxiv.org/abs/2006.11371}

\bibitem{david1993ExpertSystems}
  David JM., Krivine JP., Simmons R.: Second Generation Expert Systems: A Step Forward in Knowledge Engineering. In: David JM., Krivine JP., Simmons R. (eds) Second Generation Expert Systems. Springer, Berlin, Heidelberg (1993). \url{https://doi.org/10.1007/978-3-642-77927-5_1}

\bibitem{doanPhysicsESN}
  Doan N.A.K., Polifke W., Magri L.: Physics-Informed Echo State Networks for Chaotic Systems Forecasting. In: Rodrigues J. et al. (eds) Computational Science -- ICCS 2019. ICCS 2019. Lecture Notes in Computer Science, vol 11539. Springer, Cham. (2019). \url{https://doi.org/10.1007/978-3-030-22747-0_15}

\bibitem{europeGDPR}
  General Data Protection Regulation. \url{https://gdpr-info.eu/}

\bibitem{intelLabsNeuromorphic}
  Intel Labs: Neuromorphic Computing - Next Generation of AI. \url{https://www.intel.com/content/www/us/en/research/neuromorphic-computing.html}

\bibitem{catastrophicInfe}
  Robert M. French: Catastrophic forgetting in connectionist networks. Trends in Cognitive Sciences, 3, issue 4, 128--135 (1999). \url{https://doi.org/10.1016/S1364-6613(99)01294-2}

\bibitem{GarbinDropout2020}
  Christian Garbin, Xingquan Zhu, Oge Marques: Dropout vs. batch normalization: an empirical study
of their impact to deep learning. Multimed Tools Appl 79, 12777--12815 (2020). \url{https://doi.org/10.1007/s11042-019-08453-9}

\bibitem{Goodfellow2015ExplainHarnessAdversarial}
  Ian Goodfellow, Jonathon Shlens and Christian Szegedy: Explaining and Harnessing Adversarial Examples. International Conference on Learning Representations (2015). \url{http://arxiv.org/abs/1412.6572}

\bibitem{GrossiParametersTimeSeries}
  Grossi E.: How artificial intelligence tools can be used to assess individual patient risk in cardiovascular disease: problems with the current methods. BMC Cardiovasc Disord 6, 20 (2006). \url{https://doi.org/10.1186/1471-2261-6-20}

\bibitem{jiangESNspectralRad}
  Jiang J, Lai Y.-C.: Model-free prediction of spatiotemporal dynamical systems with recurrent neural networks: Role of network spectral radius. Phys. Rev. Research, Volume 1, Issue 3, 033056-1--033056-14, American Physical Society, (2019). \url{https://link.aps.org/doi/10.1103/PhysRevResearch.1.033056}
  
\bibitem{Ilahi2020AdversarialAtt}
  Inaam Ilahi et al.: Challenges and Countermeasures for Adversarial Attacks on Deep Reinforcement Learning. arXiv e-print (2020). \url{https://arxiv.org/abs/2001.09684}

\bibitem{Karpathy2017SW}
  Andrej Karpathy: Software 2.0. medium.com (2017). \url{https://medium.com/@karpathy/software-2-0-a64152b37c35}

\bibitem{MensahCapsuleNetworks2019}
  Mensah Kwabena Patrick, Adebayo Felix Adekoya, Ayidzoe Abra Mighty, Baagyire Y. Edward: Capsule Networks -- A survey. Journal of King Saud University -- Computer and Information Sciences, 1319--1578 (2019). \url{https://doi.org/10.1016/j.jksuci.2019.09.014}
  
\bibitem{Montavon2019LRP}
  Montavon G., Binder A., Lapuschkin S., Samek W., Müller KR.: Layer-Wise Relevance Propagation: An Overview. In: Samek W., Montavon G., Vedaldi A., Hansen L., Müller KR. (eds) Explainable AI: Interpreting, Explaining and Visualizing Deep Learning. Lecture Notes in Computer Science, vol 11700. Springer, Cham. (2019). \url{https://doi.org/10.1007/978-3-030-28954-6_10}

\bibitem{pathakAttractors2017}
  Jaideep Pathak et al.: Using machine learning to replicate chaotic attractors and calculate Lyapunov exponents from data. Chaos 27, 121102 (2017). \url{https://doi.org/10.1063/1.5010300}

  
\bibitem{Raffin2019DecouplingFeatExtr}
  Antonin Raffin, Ashley Hill, René Traoré, Timothée Lesort, Natalia Díaz-Rodríguez, David Filliat: Decoupling feature extraction from policy learning: assessing benefits of state representation learning in goal based robotics. In SPiRL Workshop ICLR (2019). \url{https://openreview.net/forum?id=Hkl-di09FQ}

\bibitem{RichterLSTMblog}
  Julius Richter: Machine Learning Approaches for Time Series. Posted on dida.do (2020). \url{https://dida.do/blog/machine-learning-approaches-for-time-series}
  
\bibitem{Singh2020ExplainableDeepLearning}
  Amitojdeep Singh, Sourya Sengupta, Vasudevan Lakshminarayanan: Explainable Deep Learning Models in Medical Image Analysis. J. Imaging 2020, 6(6), 52 (2020). \url{https://doi.org/10.3390/jimaging6060052}
  
\bibitem{Strehlitz2019InterviewTrapp}
  Markus Strehlitz: \foreignlanguage{ngerman}{Wir können keine Garantien für das Funktionieren von KI geben}. Interview with Prof. Dr. habil. Mario Trapp, director of Fraunhofer IKS (2019). \url{https://barrytown.blog/2019/06/25/wir-koennen-keine-garantien-fuer-das-funktionieren-von-ki-geben/}

\bibitem{SuOnePixelAttack2019}
  Su J., Vargas DV. and Sakurai K.: One Pixel Attack for Fooling Deep Neural Networks, IEEE Transactions on Evolutionary Computation, vol. 23, no. 5, pp. 828--841 (2019). \url{http://doi.org/10.1109/TEVC.2019.2890858}

\bibitem{Tang2020NetworksChaos}
  Yang Tang, Jürgen Kurths, Wei Lin, Edward Ott, Ljupco Kocarev: Introduction to Focus Issue: When machine learning meets complex systems: Networks, chaos, and nonlinear dynamics. Chaos 30, 063151 (2020). \url{https://doi.org/10.1063/5.0016505}

  
\bibitem{TricentisAIApproaches}
 Tricentis: AI In Software Testing. AI Approaches Compared: Rule-Based Testing vs. Learning. \url{https://www.tricentis.com/artificial-intelligence-software-testing/ai-approaches-rule-based-testing-vs-learning/}

\bibitem{verzelliCriticality2019}
  Pietro Verzelli, Cesare Alippi, Lorenzo Livi: Echo State Networks with Self-Normalizing Activations on the Hyper-Sphere. Sci Rep 9, 13887 (2019). \url{https://doi.org/10.1038/s41598-019-50158-4}

 
\bibitem{VoigtInferenceEteroclinicNet}
   Maximilian Voit and Hildegard Meyer-Ortmanns: Dynamical Inference of Simple Heteroclinic Networks: Frontiers in Applied Mathematics and Statistics (2019). \url{https://doi.org/10.3389/fams.2019.00063}
 
\bibitem{zhangInterpretableCNN}
Zhang Q., Wu YN., Zhu S.: Interpretable Convolutional Neural Networks. 2018 IEEE/CVF Conference on Computer Vision and Pattern Recognition, Salt Lake City, UT, USA, 8827--8836 (2018). \url{https://doi.org/10.1109/CVPR.2018.00920}
 
\end{thebibliography}
\end{document}